# SBCFormer: Lightweight Network Capable of Full-size ImageNet Classification at 1 FPS on Single Board Computers

Xiangyong Lu[1]    Masanori Suganuma[1,2]    Takayuki Okatani[1,2]
[1]Graduate School of Information Sciences, Tohoku University    [2]RIKEN Center for AIP
{xylu,suganuma,okatani}@vision.is.tohoku.ac.jp

**Abstract**

*Computer vision has become increasingly prevalent in solving real-world problems across diverse domains, including smart agriculture, fishery, and livestock management. These applications may not require processing many image frames per second, leading practitioners to use single board computers (SBCs). Although many lightweight networks have been developed for "mobile/edge" devices, they primarily target smartphones with more powerful processors and not SBCs with the low-end CPUs. This paper introduces a CNN-ViT hybrid network called SBCFormer, which achieves high accuracy and fast computation on such low-end CPUs. The hardware constraints of these CPUs make the Transformer's attention mechanism preferable to convolution. However, using attention on low-end CPUs presents a challenge: high-resolution internal feature maps demand excessive computational resources, but reducing their resolution results in the loss of local image details. SBCFormer introduces an architectural design to address this issue. As a result, SBCFormer achieves the highest trade-off between accuracy and speed on a Raspberry Pi 4 Model B with an ARM-Cortex A72 CPU. For the first time, it achieves an ImageNet-1K top-1 accuracy of around 80% at a speed of 1.0 frame/sec on the SBC. Code is available at https://github.com/xyongLu/SBCFormer.*

## 1. Introduction

Deep neural networks have been used in various computer vision tasks across different settings, which require running them for inference on diverse hardware. To meet this demand, numerous designs of deep neural networks have been proposed for mobile and edge devices. Since the introduction of MobileNet [27], many researchers have proposed various architectural designs of convolutional neural networks (CNNs) for mobile devices [46, 49, 68]. Moreover, following the introduction of the vision transformer (ViT) [12], several attempts have been made to adapt ViT for mobile devices [4, 8, 42, 65]. The current trend involves developing CNN-ViT hybrid models [20, 21, 35, 50]. Thanks to these studies, while ViTs were previously considered slow and lightweight CNNs were the only viable option for mobile devices, recent hybrid models for mobile devices surpass CNNs in the trade-off between computational efficiency and inference accuracy [14, 31, 32, 44].

Previous studies have mainly focused on smartphones as "mobile/edge" devices. Although processors in smartphones are less powerful than GPUs/TPUs found in servers, they are still quite powerful and would be considered in the mid-range on the spectrum of processors. There are "low-end" processors such as CPUs/MPUs for embedded systems, which usually have by far limited computational power. Nonetheless, these processors have been utilized in various real-world applications such as smart agriculture [41, 69] and AI applications for fishery [60] and livestock management [2, 30], where limited computational resources are sufficient. For example, in object detection to prevent damage by wild animals, processing dozens of frames per second may not be necessary [1]. In many cases, processing at around one frame per second is practical. In fact, lightweight models, such as MobileNet and YOLO, have been quite popular in such applications, often implemented using a camera-equipped single board computer (SBC).

This study focuses on low-end processors, which have been underexplored in the development of lightweight networks. Given their constraints, we introduce an architectural design named SBCFormer. A central question guiding our research is the suitability of either convolution or the Transformer's attention mechanism for SBCs. As outlined in [14], convolution requires complex memory access patterns, necessitating high IO throughput for efficient processing, whereas attention is comparatively simpler. Additionally, both are translated to matrix multiplications, and attention usually deals with smaller matrix dimensions compared to the traditional im2col convolution approach.

Considering that SBCs are inferior to GPUs in parallel computation resources and memory bandwidth, attention emerges as the preferred foundational building block for SBCs. Nonetheless, the attention computation carries a computational complexity that's quadratic to the number of tokens. Thus, it's crucial to maintain a low spatial resolution in feature maps to ensure computational efficiency and reduced latency. (Note that a feature map with a spatial resolution of $H \times W$ corresponds to $HW$ tokens.)

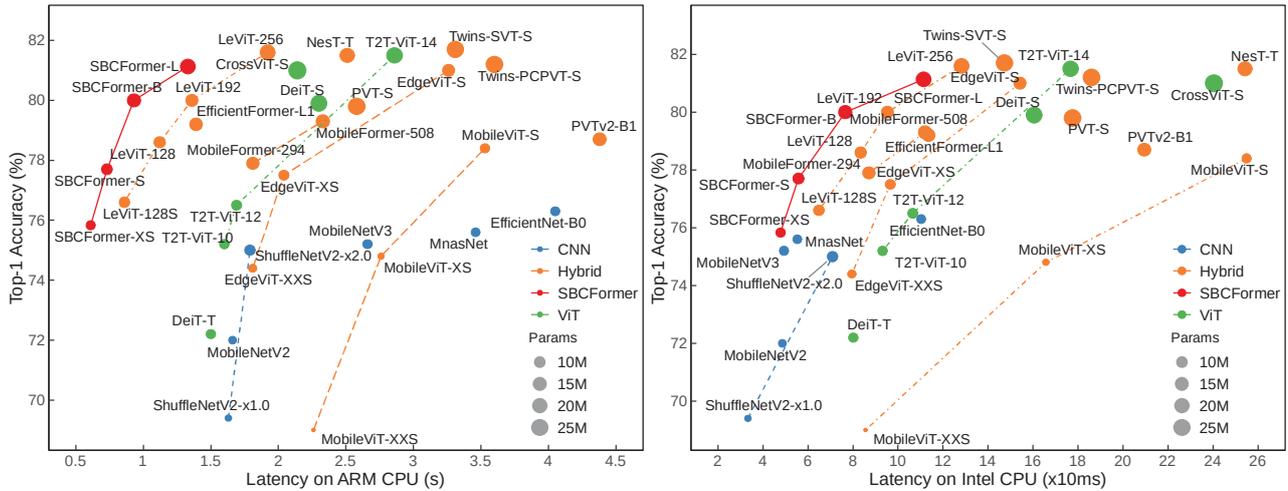

Figure 1. **Accuracy vs. Latency**. All models are trained on ImageNet-1K classification and measured the latency performance on ARM and Intel CPUs.

Using the ViT architecture, which keeps consistent resolution feature maps across all layers, leads to a loss of local details from the input image because of the coarse feature maps. In response, recent models aiming for computational efficiency, especially CNN-ViT hybrids [32, 40, 42, 54], adopt a foundation more like CNNs. In these models, feature maps reduce their spatial resolutions via downsampling from input to output. Given that applying attention to all layers can greatly increase computational costs, especially in layers with high spatial resolutions, these models use attention mechanisms only in the upper layers. This design takes advantage of the Transformer's attention mechanism, known for its strength in global interaction of image features, while retaining local details in the feature maps. However, for SBCs, convolutions in the lower layers might become problematic, causing longer computational times.

To tackle the challenge of preserving local information while optimizing attention computation, our SBCFormer employs a two-stream block structure. The first stream shrinks the input feature map, applies attention to the reduced number of tokens, and then reverts the map to its initial size, ensuring efficient attention computation. Recognizing the potential loss of local information from downsizing, the second stream acts as a 'pass-through' to retain local information in the input feature map. These streams converge, generating a feature map enriched with both local and global information, primed for the subsequent layer. Furthermore, we have refined the Transformer's attention mechanism to offset any diminished representational capacity from concentrating on smaller feature maps.

Our experiments demonstrate the effectiveness of SBC-Former; see Fig. 1. As a result of the advancements mentioned above, SBCFormer achieves the highest accuracy-speed trade-off on a widely used single board computer (SBC), namely a Raspberry Pi 4 Model B with an ARM Cortex-A72 CPU. In fact, SBCFormer attains an ImageNet-1K top-1 accuracy close to 80.0% at a speed of 1.0 frame per second on the SBC, marking the first time this level of performance has been achieved.

## 2. Related Work

### 2.1. Convolutional Networks for Mobile Devices

In recent years, there has been a growing demand for deep neural networks in vision applications across various fields, urging researchers to pay their attention towards efficient neural network design. One approach involves making convolutions computationally more efficient, as demonstrated by works like SqueezeNet [28] and so on. MobileNet [27] introduces depth-wise separable convolutions to alleviate the expensive computational cost of a standard convolutional layer, to meet the resource constraints of edge devices. MobileNetV2 [46] improves the design, introducing the inverted residual block. Our proposed SBCFormer employs the block as a primary building block for convolutional operation.

Another approach aimed at efficient designs of convolutional neural networks (CNN) architectures, as demon-strated in works such as Inception [47] and MnasNet [48]. Other studies have proposed lightweight models, including ShuffleNetv1 [68], ESPNetv2 [43], GhostNet [17], MobileNeXt [71], EfficientNet [49], and TinyNet [18], among others.

It is worth noting that CNNs, including those previously mentioned, can only capture local spatial correlations in images at each layer and do not account for global interactions. Another important point to consider is that convolution with standard-sized images can be computationally expensive for CPUs since it requires large matrix multiplications. These are areas where Vision Transformer (ViT) [12].

### 2.2. ViTs and CNN-ViT Hybrids for Mobile Devices

Thanks to the self-attention mechanism [56] and large-scale image datasets, Vision Transformer (ViT) [12] and related ViT-based architectures [3, 6, 29, 52, 72] have attained state-of-the-art inference accuracy in various visual recognition tasks [16, 67]. Nevertheless, to fully leverage their potential, ViT-based models typically require significant computational and memory resources, which have limited their deployment on edge devices that have resource constraints. Subsequently, a series of studies have focused on enhancing the efficiency of ViTs from various perspectives. Inspired by hierarchical designs in convolutional architectures, some works have developed new architectures for ViTs [24, 58, 62, 66]. Neural architecture search methods have been also utilized to optimize ViT-based architectures [7, 13]. Additionally, to reduce the computational complexity of ViTs, some researchers have proposed efficient self-attention mechanisms [5, 15, 25, 57], while others focus on utilizing new parameter efficiency strategies [23, 51].

Subsequent studies have demonstrated that incorporating convolutions in Transformer blocks can improve both the performance and efficiency of ViT-based models. For example, LeViT [14] reintroduces a convolutional stem at the beginning of the network to learn low-resolution features, rather than using the patchy stem in ViT [12]. EdgeViT [44] introduces Local-Global-Local blocks to better integrate self-attention and convolution, allowing the model to capture spatial tokens with varying ranges and exchange information between them. MobileFormer [8] parallelizes MobileNet and Transformer to encode both local and global features and fuses the two branches through a bidirectional bridge. MobileViT [42] treats Transformer blocks as convolutions and develops a MobileViT block to effectively learn both local and global information. Finally, EfficientFormer [32] employs a hybrid approach that combines convolutional layers and self-attention layers to achieve a balance between accuracy and efficiency.

Despite active research in developing hybrid models for mobile devices, there are still several issues that need to be addressed. Firstly, many of these studies do not use latency (i.e., inference time) as the primary metric for evaluating efficiency, which will be discussed later. Secondly, low-end CPUs are often excluded from these studies, with the targets limited to smartphones' CPUs/NPUs and Intel CPUs at best. LeViT, for example, was evaluated on an ARM CPU, specifically the ARM Graviton 2, which is designed for cloud servers.

### 2.3. Evaluation Metrics

There are multiple metrics for assessing computational efficiency, including the number of model parameters, operations (i.e., FLOPs), inference time or latency, and memory usage. While all of these metrics are relevant, latency is of particular interest in the context of this study. It is noteworthy that Dehghani et al. [10] and Vasu et al. [55] show that efficiency in terms of latency does not correlate well with the number of FLOPs and parameters.

As mentioned earlier, several studies have focused on developing lightweight and efficient CNNs. However, only a handful of them, such as MNASNet [48], MobileNetv3 [26], and ShuffleNetv2 [39], have directly optimized for latency. The same holds true for studies on CNN-ViT hybrids, where some are primarily designed for mobile devices [8, 42]; most of these studies did not prioritize latency as a target but instead focused on metrics like FLOPs [8].

Latency is often avoided in such studies for a good reason. It is because the instruction set of each processor and the compilers used with it heavily influence latency. Therefore, obtaining practical evaluation results necessitates choosing specific processors at the expense of general discussion. In this paper, we select CPUs used in a single board computer, such as Raspberry Pi, as our primary target, which is widely employed in various fields for edge applications. It is equipped with a microprocessor, ARM Cortex-A72, specifically designed for mobile platforms as part of the ARM Cortex-A series.

## 3. CNN-ViT Hybrid for Low-end CPUs

We aim to develop a CPU-friendly ViT-CNN hybrid network that achieves a better trade off between test-time latency and inference accuracy.

### 3.1. Principle of Design

We adopt the fundamental architecture that is commonly used in recent CNN-ViT hybrids. The network's initial stage comprises a set of standard convolution layers, which excel at converting the input image into a feature map [14, 32, 63], rather than a linear mapping from image patches to tokens in ViT [12, 51]. The main section of the network is divided into multiple stages, and the feature maps are reduced in size between consecutive stages. This results in a pyramid structure of feature maps with dimensions of $H/8 \times W/8$, $H/16 \times W/16$, $H/32 \times W/32$, and so on.

The computational complexity of the Transformer attention mechanism increases quadratically with the number of tokens, i.e., the size $h \times w$ of the input feature map. Then, the lower stages with larger-sized feature maps need more computational cost. Some studies have addressed this issue by applying attention only to sub-regions/tokens of the feature maps [19, 36, 45, 53]. Studies targeting mobile devices typically adopt attention only in high layers [32, 40, 54]. While avoiding the increased computational cost, this leads to suboptimal inference accuracy as it gives up on one of the most important properties of ViT, i.e., aggregating global information in images.

Taking these considerations into account, we propose a method of downsizing the input feature map, applying attention to the downsized feature map, and then upsizing the resulting feature map. In our experiments, we downsized the feature map to $7 \times 7$ regardless of the stage, for an input image of size $224 \times 224$. This hourglass design allows us to aggregate global information from the entire image while minimizing computational costs.

However, downsizing the feature map to this small size can lead to a loss of local information. To address this issue, we design a block with two parallel streams: one for local features and the other for global features. Specifically, we maintain the original feature map size for the local stream and do not perform an attention operation. For the global stream, we employ the above hourglass design of attention, which first downsizes the feature map, applies attention, and then upsizes it to the original size. The outputs from the two streams are merged and transferred to the next block. More details are given in Sec. 3.3. Additionally, to compensate for the loss of representational power due to the hourglass design, we propose a modified attention mechanism. See Sec. 3.4.

### 3.2. Overall Design of SBCFormer

Figure 2 shows the overall architecture of the proposed SBCFormer. The network begins with an initial section (labeled as 'Stem' in the diagram) that consists of three convolution layers with $3 \times 3$ kernels and stride = 2, which converts an input image into a feature map. The main section comprises three stages, each of which is connected to the next stage by a single convolution layer (labeled as 'Embedding' in the diagram). This layer uses a stride-two $3 \times 3$ convolution to halve the size of the input feature map. Regarding the output section, we employ global average pooling followed by a fully-connected linear layer for the final layer of the network, specifically for image classification tasks.

### 3.3. SBCFormer Block

We denote the input feature map to the block at $i$-th stage by $\mathbf{X}_i \in \mathbb{R}^{(H/2^{i+2}) \times (W/2^{i+2}) \times C_i}$.

To start the block, we place $m_i$ consecutive inverted residual blocks [46], which is first used in MobileNetV2 [46]. We use a variant with a GeLU activation function, which consists of a point-wise convolution, a GeLU activation function, and a depth-wise convolution with a $3 \times 3$ filter. We call this *InvRes* in what follows. They convert the input map $\mathbf{X}_i$ into $\mathbf{X}_i^l$ as

$$\mathbf{X}_i^l = \mathcal{F}_{\text{InvRes}}^{m_i}(\mathbf{X}_i), \tag{1}$$

where $\mathcal{F}^{m_i}(\cdot)$ indicates the application of $m_i$ consecutive InvRes blocks to the input.

As shown in Fig. 2, the updated feature $\mathbf{X}_i^l$ is transferred to two different branches, local and global streams. For the local stream, $\mathbf{X}_i^l$ is passed through to the end section of the block. For the global stream, $\mathbf{X}_i^l$ is first downsized to $h \times w$ by an average pooling layer, denoted as 'Pool' in Fig. 2. We set it to $7 \times 7$ regardless of stages in our experiments. The downsized map is then passed to a block consisting of two consecutive InvRes blocks, denoted as 'Mixer' in the diagram and next to a stack of attention blocks named 'MAttn.' The output feature map is then upsized followed by convolution, which is denoted by 'ConvT.' These operations provide a feature map $\mathbf{X}_i^g \in \mathbb{R}^{(H/2^{i+2}) \times (W/2^{i+2}) \times C_i}$ as

$$\mathbf{X}_i^g = \text{ConvT}\left[\mathcal{F}_{\text{MAttn}}^{L_i}[\text{Mixer}[\text{Pool}(\mathbf{X}_i^l)]]\right], \tag{2}$$

where $\mathcal{F}_{\text{MAttn}}^{L_i}(\cdot)$ indicates the application of $L_i$ consecutive MAttn blocks to the input.

In the last section of the block, the local stream feature $\mathbf{X}_i^l$ and global stream feature $\mathbf{X}_i^g$ are fused to obtain a new feature map, as shown in Fig. 2. To fuse the two, we first modulate $\mathbf{X}_i^l$ with a weight map created from $\mathbf{X}_i^g$. Specifically, we compute $\mathbf{W}_i^g \in \mathbb{R}^{(H/2^{i+2}) \times (W/2^{i+2}) \times C_i}$ as

$$\mathbf{W}_i^g = \text{Sigmoid}\left[\text{Proj}(\mathbf{X}_i^g)\right], \tag{3}$$

where Proj indicates a point-wise convolution followed by batch normalization. We then multiply it to $\mathbf{X}_i^l$ and concatenate the resulting map with $\mathbf{X}_i^g$ in the channel dimension as

$$\mathbf{X}_i^u = [\mathbf{X}_i^l \odot \mathbf{W}_i^g, \mathbf{X}_i^g] \tag{4}$$

where $\odot$ is the Hadamard product. Finally, the fused feature $\mathbf{X}_i^u$ is passed through another projection block to halve the channels, yielding the output of this block.

### 3.4. Modified Attention Mechanism

The above two-stream design will compensate for the loss of local information caused by the proposed hourglass attention computation. However, as the attention operates on a very low-resolution (or equivalently, small-sized) feature map, the attention computation itself must lose its representational power. To compensate for the loss, we make a few modifications to the Transformer attention mechanism; see 'MAttn' in Fig. 2.

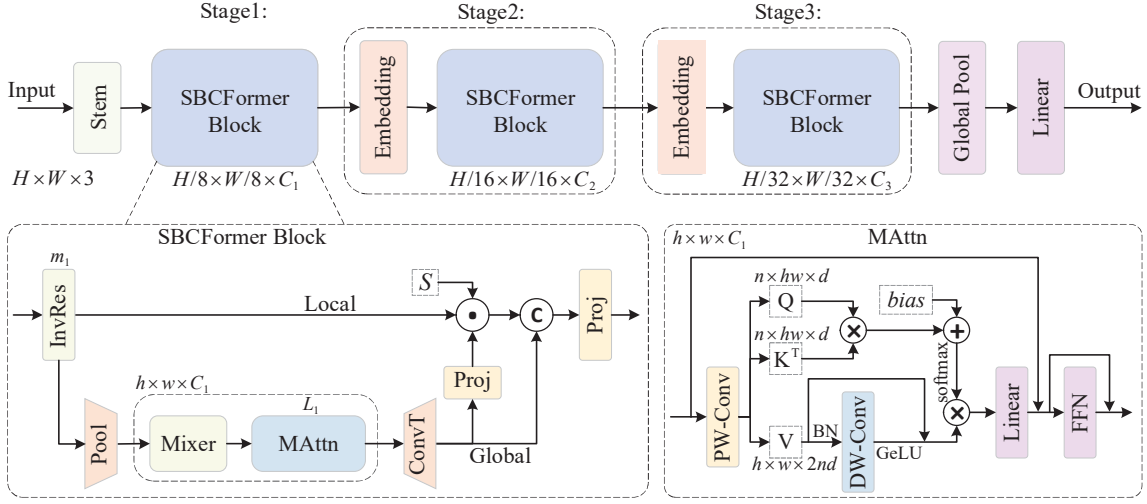

Figure 2. **Overview of SBCFormer**. The architecture of the network is hierarchical. See Sec. 3.2 for the overall design, Sec. 3.3 for the SBCFormer block, and Sec. 3.4 for the modified attention mechanism ('MAttn' in the diagram), respectively.

The main concept is to utilize the standard computation tuple of a CNN for an input to attention, specifically a $3 \times 3$ (depth-wise) convolution, a GeLU activation function, and batch normalization. The input to attention is composed of query, key, and value, and we apply the tuple to value since it forms the basis for the attention's output. Our objective is to enhance the representational power by facilitating the aggregation of spatial information in the input feature map, while simultaneously reducing the training difficulty. To offset the increase in computational cost, we eliminate the independent linear transformations applied to query and key and instead apply the identical point-wise convolution to all three components.

The details of the modified attention computation are as follows. Let $\mathbf{X} \in \mathbb{R}^{h \times w \times C_i}$ denote the input to the attention mechanism. The output $\mathbf{X}'' \in \mathbb{R}^{h \times w \times C_i}$ is computed as

$$\mathbf{X}'' = \mathrm{FFN}(\mathbf{X}') + \mathbf{X}', \quad (5)$$

where FFN stands for a feed-forward network as in ViTs [12, 51] and $\mathbf{X}'$ is defined to be

$$\mathbf{X}' = \mathrm{Linear}(\mathrm{MHSA}(\mathrm{PW\text{-}Conv}(\mathbf{X}))) + \mathbf{X}, \quad (6)$$

where Linear is a linear layer with learnable weights and PW-Conv indicates a point-wise convolution; MHSA is defined by

$$\mathrm{MHSA}(\mathbf{Y}) = \mathrm{Softmax}\left(\frac{\mathbf{Y} \cdot \mathbf{Y}^\top}{\sqrt{d}} + \mathbf{b} \cdot \mathbf{1}^\top\right) \cdot \mathbf{Y}', \quad (7)$$

where $d$ is the channel number of each head in query and key; $\mathbf{b} \in \mathbb{R}^{hw}$ is a learnable bias acting as positional encoding [14, 32]; $\mathbf{1} \in \mathbb{R}^{hw}$ is an all-one vector; $\mathbf{Y}'$ is defined by

$$\mathbf{Y}' = \mathrm{DW\text{-}Conv}_G(\mathrm{BN}(\mathbf{Y})) + \mathbf{Y}, \quad (8)$$

where $\mathrm{DW\text{-}Conv}_G$ indicates a depth-wise convolution followed by GeLU and BN is batch normalization applied in the same way as that in CNNs.

## 4. Experimental Results

We conduct experiments to evaluate SBCFormer and compare it with existing networks on two tasks, image classification using ImageNet1K [11] and object detection using the COCO dataset [34].

### 4.1. Experimental Settings

SBCFormer primarily targets low-end CPUs that are commonly used in single board computers. Additionally, we evaluate its performance on an Intel CPU commonly found in edge devices, as well as on a GPU used in desktop PCs. We use the following three processors and platforms for our experiments.

- An ARM Cortex-A72 processor running at 1.5 GHz on a single board computer, Raspberry PI 4 model B. While it is classified as low-end, ARM Cortex-A72 is a quad-core 64-bit processor that supports the ARM Neon instruction set. We used the 32-bit Raspberry Pi OS and PyTorch ver 1.6.0 to run networks.

- An Intel Core i7-3520M processor running at 2.9 GHz. It is a dual-core processor that is commonly used in

Table 1. Design of SBCFormer variants with different model sizes.

| Stage | Type | Resolution | Block | SBCFormer | | | |
|---|---|---|---|---|---|---|---|
| | | | | XS | S | B | L |
| stem | Patch Embed. | H/8 × W/8 | Embedding | × 3 (k = 3 × 3, s = 2) | | | |
| | | | | dim. 96 | dim. 96 | dim. 128 | dim. 192 |
| 1 | SBCFormer Block | H/8 × W/8 | InvRes | × 2 | × 2 | × 2 | × 2 |
| | | H/32 × W/32 | Mixer | × 1 | × 1 | × 1 | × 1 |
| | | | MAttn | × 2 | × 2 | × 2 | × 2 |
| 2 | Patch Embed. | H/16 × W/16 | Embedding | × 1 (k = 3 × 3, s = 2) | | | |
| | | | | dim. 160 | dim. 192 | dim. 256 | dim. 288 |
| | SBCFormer Block | H/16 × W/16 | InvRes | × 2 | × 2 | × 2 | × 2 |
| | | H/32 × W/32 | Mixer | × 1 | × 1 | × 1 | × 1 |
| | | | MAttn | × 3 | × 4 | × 4 | × 4 |
| 3 | Patch Embed. | H/32 × W/32 | Embedding | × 1 (k = 3 × 3, s = 2) | | | |
| | | | | dim. 288 | dim. 320 | dim. 384 | dim. 384 |
| | SBCFormer Block | H/32 × W/32 | InvRes | × 1 | × 1 | × 1 | × 1 |
| | | H/32 × W/32 | Mixer | × 1 | × 1 | × 1 | × 1 |
| | | | MAttn | × 2 | × 3 | × 3 | × 3 |

mobile devices such as laptops and tablets. It supports a variety of instruction sets including Intel Advanced Vector Extensions (AVX) and AVX2, which provide improved performance for vector and matrix operations. We used Ubuntu ver 18.04.5 LTS and PyTorch ver 1.10.1 to run networks.

- A GeForce RTX 2080Ti on a desk-top PC equipped with an Intel Xeon CPU E5-1650 v3. We used Ubuntu 18.04.6 LTS and PyTorch ver 1.10.1.

We implemented and tested all the networks using the PyTorch framework (version 1.10) and the Timm library [61]. For each of the existing networks we compare, we employ the author's official code but a few networks[1]. We follow previous studies [32, 44] to measure the inference time (i.e., latency) required to process a single input image. Specifically, setting the batch size to 1, we recorded the clock time on each platform. To ensure accuracy, we performed 300 inferences and reported the average latency in seconds. During measurement, we terminated any irrelevant applications that could interfere with the results. All computations used 32-bit floating point numbers. Since our focus is on inference speed rather than training, we trained all networks on a GPU server with eight Nvidia 2080Ti GPUs, and then evaluated their inference time on each platform.

---

[1]Their GitHub repositories are as follows:
DeiT: https://github.com/facebookresearch/deit,
LeViT: https://github.com/facebookresearch/LeViT,
NesT: https://github.com/google-research/nested-transformer,
PVT/PVTv2: https://github.com/whai362/PVT,
T2T-ViT: https://github.com/yitu-opensource/T2T-ViT,
Twins: https://github.com/Meituan-AutoML/Twins,
EfficientFormer: https://github.com/snap-research/EfficientFormer,
EdgeViT: https://github.com/saic-fi/edgevit.

### 4.2. ImageNet-1K

We first evaluate the networks on the most standard task, image classification of ImageNet-1K.

#### 4.2.1 Training

We train SBCFormer and existing networks from scratch for 300 epochs on the training split of ImageNet-1K dataset, which consists of 1.28 million images across 1,000 classes. We consider the four variants with different model size, SBCFormer-XS, -S, -B, and -L, as shown in Table 1. All models are trained and tested at the standard resolution of $224 \times 224$.

We followed the original author's code to train the existing networks. For training SBCFormer, we used the recipe from DeiT [51], which is summarized as follows. We employed the AdamW [38] optimizer with cosine learning rate scheduling [37], and applied a linear warm-up for the first five epochs. The initial learning rate was set to $2.5 \times 10^{-4}$, and the minimum value was set to $10^{-5}$. The weight decay and momentum were set to $5 \times 10^{-2}$ and 0.9, respectively, and a batch size of 200 was used. Data augmentation techniques, including random cropping, random horizontal flipping, mixup, random erasing, and label-smoothing, were applied during training, following [44, 51]. Random cropping was applied to the input image during training to obtain an image size of $224 \times 224$ pixels, while a single center crop of the same size was used during testing.

#### 4.2.2 Results

Table 2 shows the results of variants of SBCFormer with different model sizes and the current state-of-the-art lightweight networks targeting mobile/edge devices from three categories, CNNs, ViT variants, and CNN-ViT hybrids.

Table 2. **Performance of different networks on ImageNet-1K classification.** Values are averaged over 300 runs.

| Model | Type | Params(M) | GMACs | Top-1(%) | Inference Latency | | |
|---|---|---|---|---|---|---|---|
| | | | | | GPU(ms) | Intel(ms) | ARM(s) |
| ShuffleNetV2-x1.0 [39] | CNN | 7.4 | 0.6 | 75.0 | 9.7 | 33.3 | 1.63 |
| MobileNetV2 [46] | CNN | 3.5 | 0.3 | 72.0 | 9.8 | 48.6 | 1.66 |
| MobileNetV3 [26] | CNN | 5.4 | 0.2 | 75.2 | 12.1 | 49.3 | 2.66 |
| EfficientNet-B0 [49] | CNN | 5.3 | 0.4 | 76.3 | 16.8 | 110.3 | 4.05 |
| DeiT-T [51] | ViT | 5.7 | 1.3 | 72.2 | 11.1 | 80.1 | 1.50 |
| DeiT-S [51] | ViT | 22.5 | 4.5 | 79.9 | 11.5 | 160.5 | 2.30 |
| T2T-ViT-14 [64] | ViT | 21.5 | 4.8 | 81.5 | 15.2 | 176.7 | 2.86 |
| EfficientFormer-L1 [32] | Hybrid | 12.2 | 1.2 | 79.2 | 11.6 | 113.6 | 1.39 |
| MobileFormer-294 [8] | Hybrid | 11.4 | 0.3 | 77.9 | 38.1 | 87.1 | 1.81 |
| LeViT-256 [14] | Hybrid | 18.9 | 1.1 | 81.6 | 15.8 | 128.2 | 1.92 |
| EdgeViT-XS [44] | Hybrid | 6.7 | 1.1 | 77.5 | 13.8 | 96.6 | 2.04 |
| MobileViT-XXS [42] | Hybrid | 1.3 | 0.4 | 69.0 | 15.6 | 85.6 | 2.26 |
| MobileFormer-508 [8] | Hybrid | 14.0 | 0.5 | 79.3 | 38.3 | 112.0 | 2.33 |
| NesT-T [70] | Hybrid | 17.1 | 5.8 | 81.5 | 12.1 | 254.2 | 2.51 |
| PVT-Small [58] | Hybrid | 24.5 | 3.8 | 79.8 | 20.1 | 177.6 | 2.58 |
| MobileViT-XS [42] | Hybrid | 2.3 | 0.9 | 74.8 | 15.8 | 165.7 | 2.76 |
| EdgeViT-S [44] | Hybrid | 11.1 | 1.9 | 81.0 | 21.5 | 154.2 | 3.26 |
| Twins-SVT-S [9] | Hybrid | 24.1 | 2.8 | 81.7 | 20.8 | 147.3 | 3.31 |
| MobileViT-S [42] | Hybrid | 5.6 | 1.8 | 78.4 | 16.5 | 254.9 | 3.53 |
| PVTv2-B1 [59] | Hybrid | 14.0 | 2.1 | 78.7 | 12.8 | 209.5 | 4.38 |
| **SBCFormer-XS** | Hybrid | 5.6 | 0.7 | 75.8 | 15.8 | 47.8 | **0.61** |
| **SBCFormer-S** | Hybrid | 8.5 | 0.9 | 77.7 | 18.0 | 55.8 | **0.73** |
| **SBCFormer-B** | Hybrid | 13.8 | 1.6 | 80.0 | 18.1 | 76.6 | **0.93** |
| **SBCFormer-L** | Hybrid | 18.5 | 2.7 | 81.1 | 18.2 | 111.4 | **1.33** |

It is observed that SBCFormer variants with different model sizes achieve a higher trade-off between accuracy and latency on CPUs; ses also Fig. 1. The performance gap between SBCFormer and the other models is more pronounced on ARM CPUs than on Intel CPUs. Notably, SBCFormer only achieves mediocre or inferior trade-offs on the GPU. These results are consistent with our design goal, as SBCFormer is optimized for running faster on CPUs with limited computational resources.

Additional observations from the results on CPUs shown in Fig. 1 are as follows. Firstly, the popular lightweight CNNs, such as MobileNetV2/V3 [26, 46], ShuffleNetV2 [39], and EfficientNet [49], tend to be insufficient in terms of inference accuracy. They attain relatively low accuracy at the same level of speed as the recent hybrids. This well demonstrates the difficulty of adopting convolution in CPUs.

In addition, some of the ViT-CNN hybrids that have been developed for mobile applications are slower than CNNs with similar levels of inference accuracy. Examples of such hybrids include MobileViT and EdgeViT. There are various reasons for this. Firstly, some of these hybrids use FLOPs and/or parameter size as efficiency metrics, which do not necessarily correspond to lower latency. Secondly, some hybrids are intended for the latest models of smartphones, which have more powerful CPUs/NPUs than those used in our experiments. This can result in seemingly inconsistent findings compared to prior research.

Table 3. Ablation test with respect to the two key components of SBCFormer.

| Models | Params(M) | GMACs | Top-1 acc.(%) |
|---|---|---|---|
| SBCFormer-B | 13.8 | 1.6 | 80.0 |
| w/o local stream | 13.6 | 1.5 | 78.2 |
| w/ standard attn. | 12.8 | 1.5 | 77.8 |

### 4.2.3 Ablation Test

SBCFormer introduces two novel components, namely the block design with global and local streams (Sec. 3.3) and the modified attention mechanism (Sec. 3.4). To assess their effectiveness, we conducted an ablation test. Specifically, we chose SBCFormer-B and created two ablated models. The first is SBCFormer with the local stream removed from all SBCFormer blocks, while the second replaced the modified attention mechanism with the standard Transformer attention mechanism. We trained all models for 300 epochs. Table 3 shows the results, which confirm the efficacy of both introduced components.

### 4.3. Detection and Segmentation

Besides image classification, object detection is the most popular application. Thus, we test the performance of SBCFormer on object detection. Specifically, following the standard method, we employ SBCFormer as a backbone and place task-specific architecture on top of it to build models.

#### 4.3.1 Dataset and Experimental Configurations

We employ the COCO 2017 dataset [34] for evaluation. It consists of a training set of 118,000 images and a validation set of 5,000 images.

We select a basic network for object detection, i.e., RetinaNet [33]. We integrate several backbones to RetinaNet. We select SBC-Former-B and -L and choose a few baselines having approximately the same model size, from PVT [58], PVTv2 [59], and ResNet18 [22].

We train these backbones on the ImageNet-1K dataset. For the training of RetinaNet with different backbones, we adopt the standard protocol [9, 44, 59]. The images are resized so that the shorter side is 800 pixels while ensuring that the longer side is smaller than 1333 pixels. We employ the AdamW optimizer [38] with an initial learning rate of $1 \times 10^{-4}$ and train the models for 12 epochs, with a batch size of 16. During testing, the image size is re-scaled to 800 × 800.

#### 4.3.2 Results

Table 4 shows the results. It can be seen that models with SBCFormer backbones show comparable or better performance than the baselines. As shown in Fig. 1 and Table 2, PVT, PVTv2, and ResNet18 exhibit significantly slower inference speed, which can be a bottleneck for these detectors utilizing them as backbones.

### 5. Conclusions

We have proposed a new deep network design, called SBCFormer, that achieves a favorable balance between inference accuracy and computational speed when used with low-end CPUs, commonly found in single-board computers (SBCs). These CPUs are not efficient at performing large matrix multiplications, making the Transformer's attention mechanism more attractive than CNNs. However, attention is computationally expensive when applied to large feature maps. SBCFormer mitigates this cost by first reducing the input feature map size, applying attention to the smaller map, and then restoring it to its original size. However, this approach has side effects, such as the loss of local image information and limited representation ability of small-size attention. To address these issues, we introduced two novel designs. First, we add a parallel stream to the attention computation, which passes through the input feature map, allowing it to retain local image information. Second, we enhance the attention mechanism by incorporating standard CNN components. Our experiments have shown that SBCFormer achieves a good trade-off between accuracy and speed on a popular SBC, the Raspberry-PI 4 Model B with an ARM-Cortex A72 CPU.

Table 4. Performance of different visual backbones using RetinaNet on COCO val2017 object detection. "#P(M)" shows the number of parameters in million.

| Backbone | RetinaNet 1× | | | | | | |
|---|---|---|---|---|---|---|---|
| | #P(M) | AP | $AP_{50}$ | $AP_{75}$ | $AP_S$ | $AP_M$ | $AP_L$ |
| PVTv2-B0 [59] | 13.0 | 37.2 | 57.2 | 39.5 | 23.1 | 40.4 | 49.7 |
| EdgeViT-XXS [44] | 13.1 | 38.7 | 59.0 | 41.0 | 22.4 | 42.0 | 51.6 |
| ResNet18 [22] | 21.3 | 31.8 | 49.6 | 33.6 | 16.3 | 34.3 | 43.2 |
| **SBCFormer-B** | 22.1 | 39.3 | 59.8 | 41.5 | 21.9 | 42.7 | 53.3 |
| PVT-Tiny [58] | 23.0 | 36.7 | 56.9 | 38.9 | 22.6 | 38.8 | 50.0 |
| PVTv2-B1 [59] | 23.8 | 41.2 | 61.9 | 43.9 | 25.4 | 44.5 | 54.3 |
| PVT-Small [58] | 34.2 | 40.4 | 61.3 | 43.0 | 25.0 | 42.9 | 55.7 |
| ResNet-50 [22] | 37.7 | 36.3 | 55.3 | 38.6 | 19.3 | 40.0 | 48.8 |
| **SBCFormer-L** | 26.8 | 41.1 | 62.3 | 43.3 | 24.7 | 44.3 | 56.0 |

**Limitation**: For our experiments, we selected specific processors, namely two CPUs and a GPU, and measured the latency on each of them. Although these processors are representative in their categories, different results could be obtained with other processors. Additionally, our primary metric is inference latency. It can vary depending on several factors, including code optimization, compilers, deep learning frameworks, operating systems, and more. As a result, our experimental outcomes may not be reproducible in a different environment.

**Acknowledgments**

This work was partly supported by JSPS KAKENHI Grant Number 23H00482 and 20H05952.